\documentclass[sigconf, 9pt]{acmart}
\usepackage{multirow}
\usepackage[section]{placeins}
\AtBeginDocument{%
  \providecommand\BibTeX{{%
    \normalfont B\kern-0.5em{\scshape i\kern-0.25em b}\kern-0.8em\TeX}}}

\copyrightyear{2025}
\acmYear{2025}
\setcopyright{rightsretained}
\acmConference[SenSys '25]{The 23rd ACM Conference on Embedded Networked
Sensor Systems}{May 6--9, 2025}{Irvine, CA, USA}
\acmBooktitle{The 23rd ACM Conference on Embedded Networked Sensor
Systems (SenSys '25), May 6--9, 2025, Irvine, CA, USA}
\acmDOI{10.1145/3715014.3722089}
\acmISBN{979-8-4007-1479-5/2025/05}

\makeatletter
\gdef\@copyrightpermission{
\begin{minipage}{0.3\columnwidth}
\href{https://creativecommons.org/licenses/by/4.0/}{\includegraphics[width=0.90\textwidth]{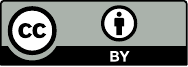}}
\end{minipage}\hfill
\begin{minipage}{0.7\columnwidth}
\href{https://creativecommons.org/licenses/by/4.0/}{This work is licensed under a Creative Commons Attribution International 4.0 License.}
\end{minipage}
\vspace{5pt}
}
\makeatother

\usepackage{enumitem}
\usepackage{xspace}
\usepackage{bm}
\usepackage{algorithmic}
\usepackage{algorithm}
\usepackage{bbm}
\usepackage{float}
\usepackage{xcolor}
\usepackage{textcomp}
\usepackage{subcaption}
\usepackage{graphicx}
\usepackage{booktabs}
\usepackage{hyperref}
\hypersetup{
    colorlinks=true,
    linkcolor=violet,
    urlcolor=cyan,
    citecolor=black,
    pdfpagemode=FullScreen,
}

\newcommand{\our}{SKELAR\xspace}

\begin{document}
\widowpenalty=10

\title{Matching Skeleton-based Activity Representations with Heterogeneous Signals for HAR}
\author{Shuheng Li}
\affiliation{
  \institution{University of California, San Diego}
  \country{}
  }
\email{shl060@ucsd.edu}

\author{Jiayun Zhang}
\affiliation{
  \institution{University of California, San Diego}
  \country{}
  }
\email{jiz069@ucsd.edu}

\author{Xiaohan Fu}
\affiliation{
  \institution{University of California, San Diego}
  \country{}
  }
\email{x5fu@ucsd.edu}

\author{Xiyuan Zhang}
\affiliation{
  \institution{University of California, San Diego}
  \country{}
  }
\email{xiyuanzh@ucsd.edu}

\author{Jingbo Shang}
\affiliation{
  \institution{University of California, San Diego}
  \country{}
  }
\email{jshang@ucsd.edu}

\author{Rajesh K. Gupta}
\affiliation{
  \institution{University of California, San Diego}
  \country{}
  }
\email{rgupta@ucsd.edu}

\renewcommand{\shortauthors}{S. Li, et al.}


\newcommand{\smallsection}[1]{\noindent\textbf{#1}}
\newcommand{\jingbo}[1]{\textcolor{blue}{\textbf{Jingbo:} #1}}
\newcommand{\jiayun}[1]{\textcolor{orange}{\textbf{Jiayun:} #1}}
\newcommand{\shl}[1]{\textcolor{cyan}{\textbf{}#1}}
\newcommand{\rev}[1]{\textcolor{black}{\textbf{}#1}}

 \newcommand{\rkg}[1]{\textcolor{red}{\textbf{Rajesh:} #1}}
 \newcommand{\todo}[1]{\textcolor{red}{\textbf{TODO:} #1}}

\begin{abstract}
    In human activity recognition (HAR), activity labels have typically been encoded in one-hot format, which has a recent shift towards using textual representations to provide contextual knowledge.
Here, we argue that HAR should be anchored to physical motion data, as motion forms the basis of activity and applies effectively across sensing systems, whereas text is inherently limited.
We propose \our, a novel HAR framework that pretrains activity representations from skeleton data and matches them with heterogeneous HAR signals.
Our method addresses two major challenges: (1) capturing core motion knowledge without context-specific details.
We achieve this through a self-supervised coarse angle reconstruction task that recovers joint rotation angles, invariant to both users and deployments;
(2) adapting the representations to downstream tasks with varying modalities and focuses. 
To address this, we introduce a self-attention matching module that dynamically prioritizes relevant body parts in a data-driven manner.
Given the lack of corresponding labels in existing skeleton data, we establish MASD, a new HAR dataset with IMU, WiFi, and skeleton, collected from 20 subjects performing 27 activities.
This is the first broadly applicable HAR dataset with time-synchronized data across three modalities.
Experiments show that \our achieves the state-of-the-art performance in both full-shot and few-shot settings.
We also demonstrate that \our can effectively leverage synthetic skeleton data to extend its use in scenarios
without skeleton collections.

\end{abstract}



\maketitle

\section{Introduction}
Human activity recognition (HAR) is a key objective in modern smart sensing systems, with applications spanning physical fitness~\cite{chadha2023human}, healthcare~\cite{gul2020patient}, and smart home~\cite{du2019novel}.
Existing works have focused on improving deep learning frameworks for heterogeneous sensing modalities, including wearable and wireless devices 
~\cite{chowdhury2022tarnet, qian2022makes, li2021two, koupai2022self, yang2022autofi}.
Despite their effectiveness, prior works overlook the physical aspects of activity labels such as the posture, movement, and trajectory. Instead, they predominantly adhere to a traditional classification scheme that employs one-hot encoding to represent activity labels. While longstanding, one-hot encoding cannot reflect the correlations within the label space.
For instance, going upstairs and downstairs are similar activities with opposite movement directions; running and walking share similar body movements but differ in frequency.
The physical characteristics of activities offer additional constraints on the prediction objective, helping to reduce model uncertainty and enhance accuracy on top of the evolution of neural network (NN) models.

Recent studies have begun to investigate textual label representations or embeddings as the target in HAR tasks~\cite{zhang2023navigating, zhang2023unleashing}. However, the semantic meaning of activity label names does not offer insight into fine-grained physical patterns like joint kinematics and body dynamics.
While providing detailed descriptions of the activity labels to a language model can enhance its understanding,
there is no
standard criterion for defining the physical meaning of human activity. 
Moreover, complex activities (e.g., cleaning the floor or playing basketball) can pose significant challenges to making precise and comprehensive textual descriptions. These limitations restrict the performance of text-based solutions.

In this paper, we argue that HAR should be anchored to physical motion data as it encodes the fundamental nature of human activity, making it broadly applicable across diverse HAR sensing systems. 
Motivated by this perspective, we propose \our, a novel HAR frameowrk that
pretrains skeleton encoders to generate activity representations and
utilizes the representations as the target to match with HAR signals coming from heterogeneous sensing systems.

\begin{figure}
\centering
    \includegraphics[width=0.46\textwidth]{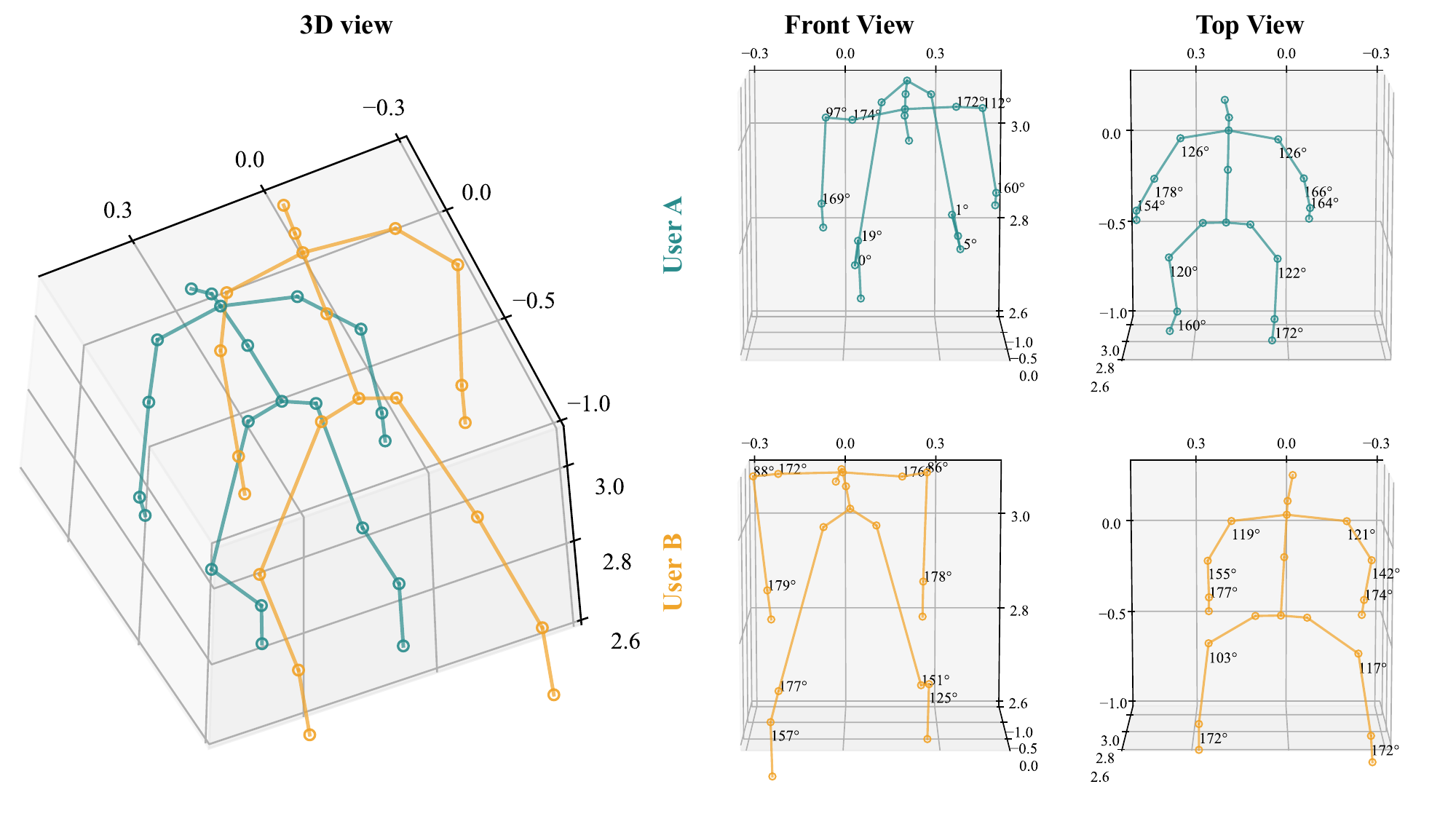}
    \caption{Skeleton data of two users in a `sitting' posture.   
    The 3D view displays their original coordinates, while the front and top views show the angles of essential limb joints. Despite the misaligned coordinates, the angles of their upper body joints are highly similar.} 
    \label{fig:skeleton_coordinates}
\end{figure}

This task faces two difficult challenges.
First, in the pretraining phase, the representation must be both informative and concise, encoding holistic motion knowledge without unnecessary user- or deployment-specific details.
Traditional representation learning relies on an auto-encoder scheme --- the encoder is trained to preserve essential features for the decoder to reconstruct the exact input.
However, this approach can lead the model to fit user-specific patterns. Figure~\ref{fig:skeleton_coordinates} demonstrates the original coordinates of the `sitting' posture performed by two different subjects. The coordinate values vary greatly due to differences in their body shapes, body positions, and locations, although the subjects maintain the same posture.
Moreover, the encoder may also fit to noise in data caused by resolution limits, hardware, object edge effect, etc. 
To address this, we propose to recover the coarse angles of the 3D rotation matrix between the edges of the body joints. This novel approach offers two significant advantages compared to coordinate reconstruction.
\begin{enumerate}[leftmargin=*]
    \item As illustrated in Figure~\ref{fig:skeleton_coordinates}, the same activities from different users have similar rotation angles. It is sufficient to infer the posture given the rotation matrices despite user and deployment variations.
    While it is possible to align the coordinates by normalizing body shapes and camera locations for each activity, doing so requires significant human effort to determine the alignment criterion for each posture frame by frame.
    
    \item Reconstructing precise angle values can also lead to fitting to noise. Given that the range of angles is naturally normalized and bounded, we propose a coarse reconstruction objective that divides the angle space into multiple identical segments. 
    The objective is converted to classify which segment the angle coarsely falls into.   
    On the other hand, adopting the same coarse objective to unbounded coordinate values is infeasible.
\end{enumerate}

Second, for the downstream application, the representation needs to be adapted to heterogeneous HAR systems with varying sensing modalities, focuses, deployment conditions, and activity label spaces.
Wearable systems capture detailed regional movement at the deployment location, while wireless systems detect general motion across the entire sensing area. Different activities also involve movements of various body parts.
To be compatible to encode full-body movements, the skeleton encoder represents an activity as a matrix using a \textbf{separate} feature vector for \textbf{each} body joint.
We accordingly propose a self-attention matching layer that computes attention scores for each body joint in a data-driven manner. It allows the downstream HAR model to focus on essential body parts that are either better sensible by the system or crucial in reflecting the physical nature of the activity. After the self-attention enhancement, the label representation is matched with features extracted by the HAR model via dot-product.

We establish a multimodal activity recognition dataset (MASD) as our testbed. Our sensing system contains wearable IMU sensors, wireless WiFi routers, and a Kinect camera to capture human motion data. To the best of our knowledge, this is the first HAR dataset with time-synchronized deployment of the three modalities.
We recruit 20 volunteers and record 27 activities of three complexity levels.
For each activity, the representation is acquired by encoding only few skeleton samples.
We also demonstrate that \our can be applied to existing systems without a Kinect camera deployed.
We explore the potential of using synthetic skeleton data to generate label representations. This is crucial when the activity cannot be traced (e.g. water sports or high-speed movements) or the sensing condition is poor (e.g. in obscured or poor-lighting environments).
We experiment on MASD, as well as two public IMU HAR datasets and two public WiFi HAR datasets.
We compare with one-hot encoding and textual label representations with ResNet and Transformer-based backbone HAR models and the experiment shows that \our consistently outperforms the compared methods.

In summary, we make the following contributions\footnote{The codes and the dataset are available at \url{https://github.com/Shuheng-Li/SKELAR}}:
\begin{itemize}[leftmargin=*]
    \item We propose \our, a novel HAR framework that first learns unified activity representations from skeleton data and matches these representation with heterogeneous sensing signals collected from various HAR systems.
    \item We propose a novel coarse angular reconstruction objective to effectively pretrain the skeleton encoder.
    \item We propose a compatible label-matching strategy that applies a self-attention mechanism to dynamically focus on essential body parts in a data-driven manner.
    \item We establish MASD, a multimodal activity sensing dataset that combines IMU, WiFi, and Kinect camera to analyze \our. We recruited 20 volunteers and recorded 27 activities of varying difficulties, presenting the first time-synchronized HAR dataset with the three modalities.\footnote{This project has been approved by IRB review.}
\end{itemize}

\section{Related Work}
\subsection{Sensor-Based Human Activity Recognition}
Sensor-based HAR is typically categorized into two types based on the sensing modality: IMU-based (wearable) and RF-based (wireless). NN models have been widely applied to these sensing systems.

\smallsection{IMU-based HAR} deploys wearable devices and utilizes motion sensors to detect human activity. Conventional deep learning-based methods focus on developing dedicated NN models, including convolutional neural networks~\cite{jeyakumar2019sensehar, dempster2020rocket, li2021units}, recurrent neural networks~\cite{ordonez2016deep},
and transformer-based models~\cite{li2021two, shavit2021boosting}.
Graph neural networks have also been applied to IMU-based HAR by viewing each sensing channel as a node in the graph~\cite{mondal2020new, liao2022deep}. However, the limited number of devices that can be deployed on the human body in daily scenarios inherently constrains the application of graph-based methods.
An alternative approach is based on self-supervised methods, which first learns time series representations through masked reconstruction before training the classification layers for downstream HAR tasks~\cite{zerveas2021transformer, liu2024focal,xu2021limu, chowdhury2022tarnet, qian2022makes}.

\smallsection{RF-based HAR} deploys radio frequency (RF) transmitters and receivers to detect user activities in the environment. Similar to IMU-based HAR, prior works have explored various NN models for a variety of modalities including WiFi Channel State Information (CSI)~\cite{chen2018wifi, xiao2020deepseg, li2021units}, radar~\cite{kim2022radar, singh2019radhar} and RF-ID~\cite{chen2023tahar}. Self-supervised methods have also gained attention in recent years, where typical self-supervised objectives are contrastive learning~\cite{hao2023bootstrapping, song2022rf, yang2022autofi}, relation prediction~\cite{saeed2021sense} and input reconstruction~\cite{koupai2022self}.

Though effective, these approaches primarily utilize traditional classification scheme that utilizes one-hot encoding to represent activity labels. This method treats activities as independent and unrelated categories, ignoring the contextual relationships inherent in human motion patterns.

\subsection{HAR with Auxiliary Information}
Parallel to the evolution of NN pipelines and architectures, recent works have begun to leverage auxiliary information to address label scarcity and incorporate contextual knowledge.

Acquiring real-world labeled activity sensing data is often heavily labor-intensive and time-consuming~\cite{leng2024imugpt}. To address this challenge, researchers have proposed using auxiliary modalities to generate virtual data.
Prior studies have explored synthesizing IMU data from various sources, including 3D motion data~\cite{kwon2020imutube, leng2024imugpt, zhang2024unimts}, audio data~\cite{liang2022audioimu} and image data~\cite{yoon2022img2imu}. Generating RF-sensing data is more difficult as it involves data with much higher dimensions. This requires the design of a dedicated pipeline that incorporates both domain expertise and advanced NN models for data generation~\cite{ahuja2021vid2doppler, bhalla2021imu2doppler}.
These works successfully enhance NN model training by alleviating the issue of label scarcity with virtual data.
While these works are not the primary focus of this paper, they offer valuable complementary insights and can be integrated into our method.

Recent advances in language models have prompted researchers to explore another direction of using text-based representations to enhance downstream classification tasks with label semantic knowledge.
Lin et al.~\cite{lin2023match} and Ranasinghe et al.~\cite{ranasinghe2023language} first propose to align video data with additional text captions in label-scarce scenarios.
This concept has also been adopted in HAR applications. SHARE proposes to train HAR models to directly generate text activity labels~\cite{zhang2023unleashing}. Furthermore, Zhang et al. expand the scope to distributed computing scenarios by utilizing text embedding space to facilitate label alignment among clients in federated learning systems~\cite{zhang2023navigating}. These methods primarily focus on text label representations. However, the semantic information of activity labels cannot fully reveal the physical meanings.

\subsection{Human Motion Representation}
Our work is also broadly related to representation learning for skeleton-based motion data. Early approaches employed an auto-encoder scheme that learns skeleton representations using generative adversarial networks~\cite{zheng2018unsupervised} or variational auto-encoders~\cite{kundu2019unsupervised}. A common limitation of these methods is that generating exact body joint coordinate values can cause the model fitting to noise and user-specific patterns. 
To address this, recent approaches have shifted towards contrastive learning, leveraging the fact that activity types remain consistent despite changes of users and camera views~\cite{li20213d, xu2022x, xu2023spatiotemporal, naimi2024rel, zhang2020video}.
While being effective for downstream skeleton-based activity recognition tasks, the learned representations do not explicitly align with other sensor-based HAR systems.
In this paper, we require the learned representation to effectively match with the features extracted by heterogeneous HAR sensing modalities. This necessitates that the skeleton representation preserves essential motion information and groups similar activities together, and avoid capturing redundant features such as the user- and deployment-specific patterns.
\section{Skeleton Representation Learning}
\begin{figure*}
    \centering
    \includegraphics[width= 0.91\linewidth]{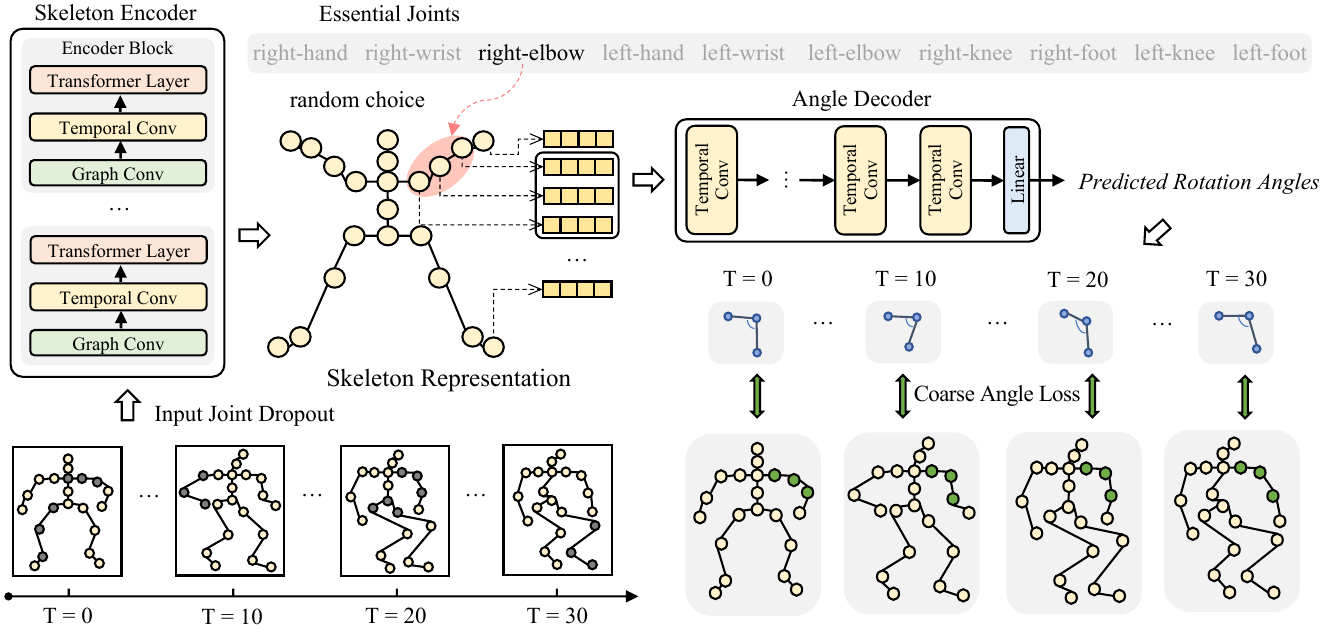}
    \caption{The pipeline of pretraining activity representations. We adopt an auto-encoder framework with a novel coarse angle reconstruction loss.
    The temporal-spatial skeleton encoder takes the input of the skeleton time series and learns a representation for each body joint. For training, the decoder picks up an essential joint from the list and recovers the 3D rotation angles using the representations of the chosen joint and its adjacent joints.}
    \label{fig:skeleton-pretrain}
\end{figure*}

As illustrated in Figure~\ref{fig:skeleton-pretrain},
we apply an auto-encoder framework with a novel coarse angle reconstruction loss to train skeleton representations.
The encoder captures both spatial and temporal information from the skeleton data and generates a separate feature vector for each body joint.
The decoder takes input from an essential joint along with its adjacent representations.
Instead of reconstructing exact angle values, the auto-encoder is designed to classify `coarse' angles, which helps mitigate overfitting to noise in the data.
We now describe these components in detail.

\subsection{Skeleton Encoder}
As shown in Figure~\ref{fig:skeleton-pretrain}, the skeleton encoder is composed of several spatio-temporal encoding blocks with various temporal resolutions. 
Let $X \in \mathbb{R}^{v \times c \times t}$ denote the input to each encoding block, where $v, c, t$ denote the number of body joints, hidden channels and time steps respectively. Particularly, we set $c = 3$ for the input of the first encoder block, to represent the 3D coordinates of the body joints. 

We first apply Decoupled Graph Convolution Neural Networks (Decoupled GCN)~\cite{cheng2020decoupling} to extract the spatial information from adjacent joints. The graph convolution is applied to the motion data of all the time steps.
Let $H \in \mathbb{R}^{v \times c}$ denote the joint features at one time step, the output $H'$ of traditional graph convolution~\cite{kipf2016semi} is:
\begin{equation}
H' = \sigma\left(\tilde{D}^{-\frac{1}{2}}A\tilde{D}^\frac{1}{2}HW\right),
\end{equation}
where $W \in \mathbb{R}^{c \times c'}$ is the trainable weight matrix, $A$ is the adjacency matrix with self-connections, and $\tilde{D}_{ii} = \sum_j\tilde{A}_{ij} +\epsilon$ represents the diagonal node degree of $\tilde{A}$ with $\epsilon = 0.0001$ to prevent empty rows. 
$\tilde{D}$ is further made trainable to improve the performance in following works~\cite{shi2019skeleton, cheng2020decoupling}.
Traditional GCN suffers from coupling issues, where all the hidden channels share the same (coupled) spatial aggregation kernel $\tilde{D}$. It is different from
temporal convolutions that have separate kernels for each input channel.
Decoupled GCN addresses this issue by explicitly dividing the hidden dimension $H$ into $g$ groups and expanding the aggregation kernel to $g$ decoupled parts. Let $H^W = HW$. Decoupled GCN is computed as:
\begin{equation}
    H' = \sigma\left(\left[
    \tilde{A}^{(1)}H^W_{[:,:\frac{H}{g}]};
    \tilde{A}^{(2)}H^W_{[:,\frac{H}{g}:\frac{2H}{g}]}; ... ;
    \tilde{A}^{(g)}H^W_{[:,\frac{(g-1)H}{g}:]}
    \right]\right).
\end{equation}
Decoupled GCN initializes each aggregation kernel $\tilde{A}^{(i)}$ identically as $\tilde{D}^{-\frac{1}{2}}A\tilde{D}^\frac{1}{2}$ but allows them to be trained independently to focus on various spatial aggregation patterns.

The temporal information is then extracted using a temporal convolution layer followed by transformer layers~\cite{vaswani2017attention}. 
The temporal convolution divides the time domain into smaller segments, reducing the time dimensions for each encoder block. The self-attention layer extracts temporal information separately for each node. 
Specifically, a single-head self-attention layer is applied separately to the hidden feature for each body joint to prevent spatial feature entanglement. We apply multiple stacks of spatio-temporal encoders to extract spatial features in various time resolutions. 

Traditional representation learning methods commonly encode $X_i$ as a single vector, summarizing both temporal and spatial features. However, the downstream HAR system may focus on different body parts depending on the sensing modality or deployment location. As a result, our skeleton encoder learns the representation for all the joints. An average pooling layer is applied to the time domain following the last spatio-temporal encoder block, resulting in $Z \in \mathbb{R}^{v \times k}$, a representation for the $v$ body joints.

\subsection{Coarse Angle Reconstruction}
Given a dataset with $N$ skeleton examples, traditional coordinate reconstruction loss trains an auto-encoder to completely recover the original input by minimizing the mean square error (MSE) loss between the input and the reconstructed output as:
\begin{equation}
\mathcal{L}_{recon} = \frac{1}{N }\sum_{i=1}^N |f_{dec}(f_{enc}(X_i)) - X_i|^2,
\label{eq:reconstruction_loss}
\end{equation}
where $f_{enc}$ and $f_{dec}$ denote the encoder and the decoder model respectively. The initial motivation of exact coordinate reconstruction is to force the encoder to preserve all the information so that the decoder can recover the original input. However, due to the noisy and heterogeneous nature of edge devices, matching the skeleton data precisely may cause the model fit to device noise and domain-specific patterns such as user body shape, user activity patterns, and device deployment location.

In our paper, we require the skeleton representation to capture motion information about body poses. It is not necessary for the representation to preserve all the information and being capable to reconstruct the original skeleton data precisely. To this consideration, we propose a novel coarse angle reconstruction task, where the auto-encoder is trained to recover the angular information of body joints in 3D space.
Intuitively, human body pose can be strictly determined if the `direction'' of all the edges (bones) in the body skeleton is fixed, while the `length' of the edges just reflects the differences in body shapes rather than body poses. As the example shown in Figure~\ref{fig:skeleton_coordinates},
the skeleton data of the same sitting posture closely resemble one another in terms of joint angles, but their coordinates in 3D view are not aligned.

Theoretically, the `direction' of the edges can be inferred from the rotation angles between the edge pairs that form body joints. 
Given 3D vector $e_1 = (x_1, y_1, z_1)$ and $e_2 = (x_2, y_2, z_2)$, the projected rotation angles $\theta = (\theta_x, \theta_y, \theta_z)$ along the $x, y, z$ axes are:
\begin{equation}
\begin{split}
    \theta_x = \arccos \frac{y_1y_2 + z_1z_2}{\sqrt{y_1^2+y_2^2}\cdot \sqrt{z_1^2+z_2^2} }
    \\
    \theta_y = \arccos \frac{x_1x_2 + z_1z_2}{\sqrt{x_1^2+x_2^2}\cdot \sqrt{z_1^2+z_2^2} }
    \\
    \theta_z = \arccos \frac{x_1x_2 + y_1y_2}{\sqrt{x_1^2+x_2^2}\cdot \sqrt{y_1^2+y_2^2} }
\end{split}
\end{equation}
Similar to coordinates reconstruction, recovering the exact angle values can also cause the model fitting to noise in data. Thus, we propose a coarse objective that divides a full angle into $2m$ equal intervals $[0, \pi/m), [\pi/m, 2\pi/m), ..., [(2m-1)\pi/m, 2\pi)$.
The reconstruction problem is then converted to coarsely classify which interval that $\theta_x$, $\theta_y$, and $\theta_z$ fall into.

\subsection{Pretraining Pipeline}
To prevent overfitting, we apply joint dropout to the input skeleton data. For the skeleton graph in each time step, we randomly sample the nodes with probability $\lambda$ and dropout the sampled nodes and their direct adjacency.
The dropout enables the skeleton encoder to learn to recover the graph connections, enhancing its ability to understand human motion. We do not explicitly train the model to recover the dropped nodes because recovering the masked parts in a time series is not deterministic, which harms the objective of learning general physical meanings from human activities.

To alleviate training difficulty, we randomly select only one essential joint from the list in Figure~\ref{fig:skeleton-pretrain}, and feed the representations of the chosen joint together with its adjacent joints to the decoder. 
We choose the limb joints as the essential joints, as their motion carries most information about the dynamics of an activity.   

Note that the joint representations already capture the spatial information about the neighbor joints through the graph convolution, the decoder only needs to recover the temporal information. 
Thus, we apply a 1D transpose convolution network as the decoder model, followed by three parallel linear classification heads to recover the coarse angles along the three axes separately.

The final training objective can be written as Eq.~\eqref{eq:final_loss}. $M_i$ denotes the random dropout for the $i$-th input data, and $p$ denotes the indices of the essential joints sampled in this batch. The decoder generates the prediction of $\theta_i^{(p)}$ the angular interval of the rotation matrix and the loss is computed using cross-entropy loss $\mathcal{L}_{CE}(\cdot, \cdot)$.
\begin{equation}
\begin{split}
    Z_i &= f_{enc}(M_iX_i) \quad i = 1\  ...\ N\\
    \mathcal{L} &= \frac{1}{N }\sum_{i=1}^N \mathcal{L}_{CE} \left( f_{dec}(Z_i^{(p)}), \theta_i^{(p)}) \right) 
\end{split}
\label{eq:final_loss}
\end{equation}

\section{Downstream HAR Pipeline}
\begin{figure}[t]
    \centering
    \includegraphics[width=\linewidth]{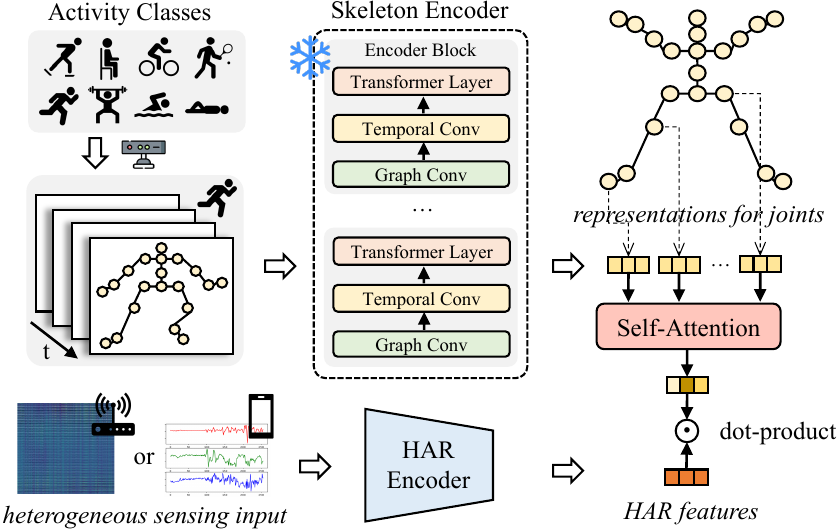}
    \caption{HAR matching pipeline. \our is compatible with various sensing modalities and backbone models. For each activity, we sample few-shot skeleton data and acquire joint representations using the pretrained encoder.
    Self-attention is applied to highlight key body parts and the predictions is made by feature matching with dot-product. }
    \label{fig:HAR_matching}
\end{figure}

Figure~\ref{fig:HAR_matching} illustrates the application of the skeleton representation in downstream HAR tasks.
Traditionally, HAR models train the encoder using the one-hot representation of labels as the target.
\our replaces the one-hot target with the objective of dynamically matching with the skeleton representation of labels.
Therefore, \our is also agnostic to the downstream encoder architectures.
While some NN structures tend to perform better with specific modalities, our experiment do not strictly adhere to the optimal choices from the recent state-of-the-art. Instead, we employ standard model architectures (e.g. ResNet and Transformer) and slightly tune layer configurations to ensure reasonable performance on each modality.
We introduce how to obtain label representations using the pretrained skeleton encoder in Section~\ref{sec:label_representation}, and the self-attention matching algorithm in Section~\ref{sec:label_matching}.

\subsection{Label Representation}
\label{sec:label_representation}
Ideally, researchers can collect representative few-shot skeleton samples from only one volunteer for each activity and feed them to the skeleton encoder.
For each activity, the representation is acquired by taking the average over these samples.
Since we only require few skeleton samples to obtain the representation, \our can be directly integrated into any existing HAR systems with minor extra skeleton data collection burdens.

Moreover, \our is flexible in adapting to shifts in label space and it can quickly incorporate unseen labels.
When a new activity is introduced, the simplest approach for adding the corresponding label is to collect skeleton motion data using compatible Kinect cameras.
This method is especially practical for large companies that provide HAR applications and need to support the classification of new activities.
The companies can continuously collect representative skeleton samples in controlled lab environments, integrating new activity representations as needed.

In cases where collecting skeleton data is not feasible due to the nature of certain activities (e.g., tracking motion data for swimming or cycling can be challenging) or device limitations, synthetic techniques can be used to generate few-shot skeleton samples.
In this paper, we demonstrate the effectiveness of using representations derived from synthetic motion data with T2MGPT~\cite{zhang2023generating}, a model that combines Vector Quantized Variational AutoEncoder (VQ-VAE) and Generative Pretrained Transformer (GPT) to synthesize human motion from textual descriptions.
We apply T2MGPT to generate two-shot skeleton motion samples for activities lacking available skeleton data. The generated samples are interpolated to match the input size required by our skeleton encoder, enabling us to derive label representations for unseen actions.

\subsection{Attention-Based Label Matching}
\label{sec:label_matching}
As shown in Figure~\ref{fig:HAR_matching}, the matching happens at the end of the pipeline, by replacing the simple linear classification layer with a dedicated matching module.

Let $Y \in \mathbb{R}^{d}$ denote the feature extracted by the HAR encoder, and $Z_i \in \mathbb{R}^{v \times k}$ denote the representation of the $i$-th activity.
Note that \our has separate feature vectors for all the body joints. The high-dimensional nature of $Z_i$ makes the matching a non-trivial problem.
An intuitive yet simple matching method is directly using dot-product to compare the similarity between $Y$ and $\bar{Z_i}$, the averaged ${Z_i}$ among the $v$ joints as:
\begin{equation}
    \text{SIM} = YW\bar{Z_i}^T, 
    \label{eq:simple_matching}
\end{equation}
where $W \in  \mathbb{R}^{d \times k}$ is a learnable parameter that matches the dimensions between $Y$ and $\bar{Z_i}$.

The simple matching assumes that all body joints contribute equally to capturing activity patterns. However, this assumption is flawed for two reasons. First, the sensing system may have varying focuses depending on the sensing modality and deployment location. Second, different activities involve the movement of distinct body parts. To address these issues, we propose an attention-based label-matching mechanism that enhances the simple matching approach by learning to prioritize key joints in a data-driven manner. Specifically, the matching module learns a self-attention mechanism to focus on the most essential body parts as:
\begin{equation}
    \begin{split}
        Q, K, V &= Z_iW_Q, Z_iW_K, Z_iW_V \\
        \text{Attention}(Q, K, V) &= \text{softmax}(\frac{QK^T}{\sqrt{d}})V \\
        Z_i' &= \sum_j^{v} \text{Attention}(Q, K, V)_{[j,:]}.
    \end{split}
\end{equation}
$W_Q, W_K, W_V \in \mathbb{R}^{k \times d}$ are the projection layers for attention query, key, and value. The attention score is calculated using self-attention computed from the query and key matrices, and the values are aggregated using the attention weights to acquire the attention-enhanced label representation $Z_i' \in \mathbb{R}^{d}$. Then we use dot-product to compute the similarity score between the HAR feature $Y$ and $Z_i'$. During training, we apply softmax to all the similarity scores between the HAR features and the label representations. During inference, the label with the highest similarity is selected as the prediction result.

\smallsection{Time Complexity.}
The self-attention matching layer does not introduce additional overhead compared to the simple matching introduced in Eq.~\eqref{eq:simple_matching}.
In our implementation, the projection layers $W_Q, W_K, W_V$ maintain the same hidden size as the input.
Therefore, the time complexity of the projection layers is $O(vdk)=O(vd^2)$, and the time complexity of the attention score calculation is $O(v^2d)$. Since $v$ represents the number of nodes in the skeleton data, this is a constant value that is typically in the range of $[20, 30]$. The time complexity can be reduced to $O(vd^2)$, which is equivalent to that of the simple matching.
After the training phase of the self-attention layer, the self-attention mechanism can be reduced to only using the cached attention-enhanced representation $Z_i'$. It further reduces the entire attention layer, making \our no additional computational cost in the inference phase.
\section{Datasets and \our Preparation}
\subsection{Skeleton Encoder Pretraining}
We use the following two large-scale human motion datasets to pretrain the skeleton encoder.
\begin{itemize}[leftmargin=*]
\item \emph{NTURGB+D-120} is the first large-scale dataset for RGB+Depth HAR,
which collects 106 subjects, containing around 114,000 video clips and more than 8 million time frames using Microsoft XBox One Kinect v2 camera~\cite{liu2019ntu}. 
This dataset collects 120 different activities with various types and we exclude all the mutual activities as we focus on learning the motion dynamics of one single user. The skeleton data contains 25 joints and we exclude 2 hand tip joints and 2 thumb joints from the data due to their lower sensing accuracy. We choose this dataset because we also deploy the same Kinect camera to collect few-shot skeleton samples for acquiring activity representations.
\item \emph{Human-ML3D} is a large captioned human motion dataset with 14,616 skeleton examples accompanied with text descriptions~\cite{guo2022generating}.
It was first introduced to facilitate training text-to-motion generation models, including T2MGPT~\cite{zhang2023generating}, the synthetic skeleton generator used in our experiments.
As a result, this dataset is an essential supplementary to NTURGB+D for it enables \our to be generalized to simulated skeleton data. The skeleton data in Human-ML3D has 22 joints with 3 spine joints in between of the neck and the middle hip, while NTURGB+D only has 2 spine joints. To align with NTURGB+D, we reduce the 3 spine joints to the 2 middle points at the edges.
\end{itemize}
We prepare the skeleton data by interpolating them to 30 Hz frequency and dividing them into 5-second chunks with 150 timestamps. We use stochastic gradient descent (SGD) with 1e-2 learning rate to optimize the skeleton encoder and decoder using the coarse angular reconstruction loss. The model is pretrained for 1,000 epochs, with the dropout ratio $\lambda$ initially set to 0 and progressively increased. Specifically, the ratio is adjusted in a stepwise manner to drop approximately 5\%, 10\%, 15\%, and 20\% of the total joints at the 200th, 400th, 600th, and 800th epochs, respectively. The output of the skeleton encoder $Z$ contains $v = 21$ joints and each joint has a separate representation vector of size $k = 256$.

\subsection{Multimodal Activity Sensing Dataset (MASD)}
\label{sec:masd}
\begin{figure}
    \centering
    \includegraphics[width=0.46\textwidth]{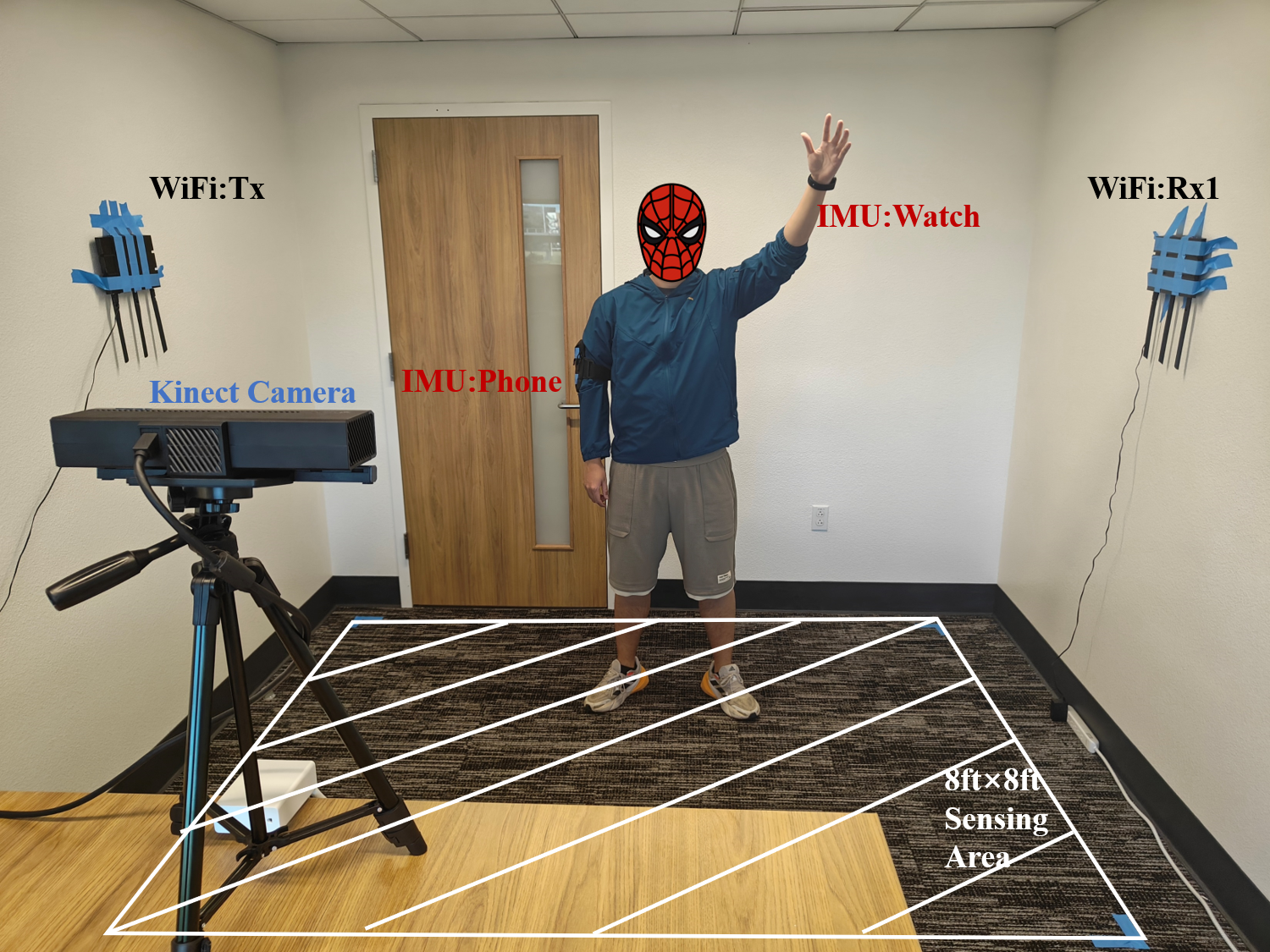}
    \caption{Sensor deployment site. Activities are performed within the 8ft $\times$ 8ft area marked by the blue tape. Routers on the wall act as the transmitter and the receiver. The volunteer wears a watch and phone with IMU sensors.}
    \label{fig:deployment}
\end{figure}

We establish a time-synchronized multimodal sensing system and collect a Multimodal Activity Sensing Dataset (MASD) to evaluate our method.
Specifically, the MASD dataset contains wearable sensors (IMU), wireless sensing (WiFi), and a Kinect camera.
We make our dataset more applicable to research in the community by simulating daily scenarios, thus choosing IMU and WiFi as they are mostly available in typical households.
\our can be potentially generalized to other wireless sensing modalities like radar and lidar. The generalization of \our is discussed in Section~\ref{sec:discussion}.
As shown in Figure~\ref{fig:deployment}, this system is deployed in an office environment. The 8ft $\times$ 8ft-sensing area is marked by the blue tape.

\emph{IMU.} To simulate daily use scenarios, we only deploy two wearable devices with IMU sensors, an Apple Watch Series 5 on the left wrist and an iPhone 10 on the right upper arm mounted with an armband. The watch is equipped with an accelerometer and a gyroscope, while the phone has a magnetometer in addition to the sensors in the watch. The sensor readings are recorded via SensorLog app\footnote{\url{https://sensorlog.berndthomas.net/}} at 30 Hz sampling rate. In our experiments, we only use the data from the accelerometer and the gyroscope.

\emph{WiFi.} We deploy two TP-Link WDR 4300 routers for WiFi CSI sensing and configure one as the transmitter and the other as the receiver. We adopt Atheros CSI toolkit~\cite{xie2015precise} to extract CSI data. The transmitter sends WiFi packets in 2.4 GHz channel with 56 subcarriers. The packet is sent at the frequency of 100 Hz.

\emph{Kinect.} We deploy a Microsoft Xbox One Kinect v2 camera facing the site. We utilize Kinect for Windows SDK 2.0\footnote{\url{https://www.microsoft.com/en-us/download/details.aspx?id=44561}}
to analyze the depth information and collect skeleton data, with a sampling frequency of 30 Hz. We use the same device as NTURGB+D and we also remove the 2 hand tip joints and 2 thumb joints.

We recruit 20 volunteers of various genders and body shapes to perform 27 activities in a relaxed and natural manner.
Each activity is performed continuously for 30 seconds. Table~\ref{tab:difficulty} summarizes all the activities collected in MASD.
In the collection phase, the camera and the routers are commanded by a laptop. Both the laptop, the smartwatch and the smartphone are connected to the campus WiFi for Network Time Protocol (NTP) synchronization. For data annotation, we directly label the skeleton data during the collection process and use the recorded timestamps as the reference to label the other two modalities. In MASD, the smartwatch data from 4 volunteers is missing due to storage issues.
During preprocessing, missing IMU timestamps are filled using interpolation, and the received CSI data is processed to extract signal amplitude.

Our preliminary experiment shows that it becomes much more difficult with the increase of activity complexity.
To ensure reasonable performance on both IMU and WiFi modalities, we create three levels of difficulties as summarized in Table~\ref{tab:difficulty}.
The difficulty level is decided based on both activity complexity and the similarity between activity pairs.
For the complexity, we define the activities that involve entire body movement (both upper body and lower body) as hard level. When performing \emph{drink water} and \emph{torso-twisting}, there is no lower body movement made by the volunteers. So they are labeled as medium level. The other activities that involve both upper body and lower body movement in Table~\ref{tab:difficulty} are labeled as hard level.
For the similarity, we intentionally design 4 confounding pairs of activities: \emph{wave right hand} and \emph{wave left hand}, \emph{kick right foot} and \emph{kick left foot}, \emph{right hand up} and \emph{left hand up}, \emph{draw clockwise} and \emph{draw counterclockwise}. They are medium activities by definition, but we only include the former one in the medium level to avoid the difficulty in distinguishing them within each pairs.

For the downstream HAR task, we experiment on three modality-label combination settings: \textbf{IMU(H)}, \textbf{IMU(M)} and \textbf{WiFi(E)}, representing IMU with hard labels, IMU with medium labels and WiFi with easy labels respectively.
We only experiment on the WiFi with easy labels for we cannot obtain reasonable performance with the chosen backbone models on the medium or the hard labels.

To acquire the representations for MASD, we randomly select a user and use the skeleton data as the representative samples. The default setting of \our samples five skeleton sequences per activity and use the average of these five samples as the label representation for each activity. We also experiment on the sensitivity to the number of skeleton data sampled in Section~\ref{sec:sensitivity}.

\subsection{Public Datasets}
\begin{table}[t]
  \centering
  \caption{Label Set Difficulty Levels}
  \small
  \begin{tabular}{ccl}\toprule
    \textbf{Level} & \textbf{\# Labels} & \textbf{Activities} \\ \midrule
    Easy & 5 & standing, walking, jumping, sitting, lying\\ \midrule 
    \multirow{3}{*}{Medium} & \multirow{3}{*}{11} & Easy activities + wave right hand, drink water \\ && torso-twisting, kick right foot, right hand up \\
    && draw clockwise\\ \midrule
    \multirow{5}{*}{Hard} & \multirow{5}{*}{27} & Medium activities + turn left, turn right, wave- \\
    && left hand, throw, kick left foot, golf swing\\
    && basketball shooting,boxing, squatting, push\\
    && pull, bending (stand), bending (sit), leg stretch\\
    &&  left hand up, draw counterclockwise \\   
    \bottomrule
  \end{tabular}
  \label{tab:difficulty}
\end{table}

\begin{table}[t]
    \centering
    \small
    \caption{Statistics of the Datasets (Tasks) for Evaluation}
    \begin{tabular}{lcccc} \toprule
      \multirow{2}{*}{Dataset} & Sampling & \multirow{2}{*}{\#Examples} & \multirow{2}{*}{\#Labels} & Sample \\
       & Rate & & & Size \\ \midrule
        MASD-IMU(H) & 30 Hz & 10180 & 27 & (150, 12)\\ 
        MASD-IMU(M) & 30 Hz & 4147 & 11 &  (150, 12)\\
        MASD-WiFi(E) & 100 Hz & 3028 & 5 & (500, 224)\\ 
        OPPORTUNITY & 33 Hz & 9599 & 4 & (256, 45)\\
        PAMAP2 & 100 Hz & 10835 & 12 & (125, 27) \\ 
        UTHAR & 1000 Hz & 8312 & 7 & (200, 90)\\ 
        NTUFI & 500 Hz & 7488 & 6 & (500, 342)\\ \bottomrule
    \end{tabular}
    \label{tab:data}
\end{table}

To expand the scope of evaluation, we also experiment on two public IMU datasets and two WiFi datasets, where labeled skeleton data is not available.
\our is flexible with unseen labels, as it can seamlessly integrate new activity representations by modeling few-shot real or synthetic samples. In this paper, we use the public datasets to demonstrate the effectiveness of \our in incorporating synthetic skeleton representations.
The detailed information of these datasets are summarized in Table~\ref{tab:data}.

\begin{itemize}[leftmargin=*]
    \item \emph{OPPORTUNITY}~\cite{roggen2010collecting} collects multi-level human activities in a room simulating a studio flat. We select a set of 9-axis IMU sensors deployed at the right lower arm, right upper arm, left lower arm, left upper arm and the back of the body. We focus on locomotion classification of \textit{standing, walking, sitting, and lying}. The data is collected continuously with a sampling rate of 33 Hz. 
    \item \emph{PAMAP2}~\cite{reiss2012introducing} collects 12 activities with 9-axis IMU sensors deployed at the chest, the dominant side wrist and the ankle. The sampling frequency of the sensors is 100 Hz.
    \item \emph{UTHAR}~\cite{yousefi2017survey} collects 7 daily activities using the Linux 802.11n CSI Tool~\cite{halperin2011tool} built on the Intel WiFi Wireless Link 5300. The sampling rate is 1000 Hz in their experiments.
    \item \emph{NTUFi}~\cite{yang2022benchmark} uses Atheros toolkit installed on TP-Link N750 routers for CSI extraction. It collects 6 daily activities and the sampling rate is 500 Hz.
\end{itemize}

\subsection{Virtual Skeleton Data Generator}
We use T2MGPT, a state-of-the-art text to motion model to generate synthetic skeleton data for the public datasets.
Experiment results have demonstrated that the motion data generated by T2MGPT is at least 95\% similar to the real motion sequences in term of R-Precisions~\cite{zhang2023generating}.   
Note that the public datasets may have some overlapped labels with MASD. However, we adopt a more difficult setting and assume the developer has no access to any skeleton data samples and \our can still improve the performance for downstream HAR. We use the template of \textit{The person is <Label Name>} as the input prompt to T2MGPT with \textit{<Label Name>} substituted by the actual label names.
The motion data generation script and the pretrained model weights are acquired in their Github repository\footnote{\url{https://github.com/Mael-zys/T2M-GPT}}.
We then adopt the same procedure as Section~\ref{sec:masd} to obtain the skeleton representations.

\section{Experiments}
\begin{table*}[t]
\caption{The Main Result on IMU Datasets. The average accuracy is reported based on five runs.}
\begin{tabular}{ccccccc}
\toprule

Model & Representation 
& MASD-IMU(H) & MASD-IMU(M) & Opportunity & PAMAP2 &  Mean \\ \midrule

\multirow{5}{*}{ResNet} 
& \our & \textbf{83.82} & \textbf{92.23} & \textbf{89.54} & \textbf{96.49} & \textbf{90.52} \\
&  One-hot & 82.09 & 90.47 & 88.43 & 96.43 & 89.35 \\
& Random  & 80.78 & 90.47 & 87.95 & 96.21 & 88.85 \\
& Llama-Emb & 80.72 & 89.18 & 84.57 & 94.24 & 87.18 \\
& Node2Vec & 83.01 & 90.60  & 87.44 & 95.40 & 89.11 \\ \midrule
\multirow{5}{*}{Transformer} 
& \our & \textbf{81.86} & \textbf{92.79} & \textbf{88.75} & \textbf{95.87} & \textbf{89.86} \\
& One-hot & 79.89 & 90.28 & 87.86 & 95.29 & 88.33 \\
& Random  & 79.62 & 90.91 & 87.75 & 95.44 & 88.43 \\
& Llama-Emb & 80.01 & 91.07 & 85.01 & 95.44 & 87.88 \\
& Node2Vec & 78.20 & 87.93 & 87.99 & 94.06 & 87.05 \\ \bottomrule
\end{tabular}
\label{tab:main_1}
\bigskip
\caption{The Main Accuracy Result on WiFi Datasets. The average accuracy is reported based on five runs.}
\begin{tabular}{cccccc}
\toprule
Model & Representation 
& MASD-WiFi(E) & NTU-Fi & UTHAR &  Mean \\ \midrule
\multirow{5}{*}{ResNet} 
& \our & \textbf{55.05} &  \textbf{93.89} & \textbf{98.38} & \textbf{82.42} \\
& One-hot & 46.97 &  92.80 & 98.01 & 79.26 \\
& Random  & 45.08 & 92.92 & 96.74 & 78.25 \\
& Llama-Emb  & 45.68 & 90.03 & 93.26 & 76.32 \\
& Node2Vec & 52.44 & 91.98 & 97.45 &  80.62 \\ \midrule
\multirow{5}{*}{Transformer} 
& \our & \textbf{65.57} & \textbf{96.46} & \textbf{98.12} & \textbf{86.72} \\
& One-hot & 59.93 & 95.73 & 96.19 & 83.95\\
& Random  & 58.96 & 94.58 & 97.74 &  83.76\\
& Llama-Emb & 56.81 & 95.96 & 97.89 & 83.55 \\
& Node2Vec  & 64.82 & 96.09 & 96.87 & 85.93 \\ \bottomrule
\end{tabular}
\label{tab:main_2}
\end{table*}

\subsection{Main Results}
\label{sec:main}
We compare the performance of \our with the following alternative label representation methods.
\begin{itemize}[leftmargin=*]
    \item \emph{One-hot.} The traditional one-hot representation.
    \item \emph{Random.} The label representation is initialized as a random vector, instead of a one-hot vector. The size of the random vector is also 256, to match the size of the representation given by our skeleton encoder.
    \item \emph{Llama-Emb.}  We obtain the embedding from Llama-2 7B model~\cite{touvron2023llama} by using a prompt-based method that utilizes the contextualized embedding of the last token~\cite{jiang2023scaling}. This method has shown great sentence embedding performance without fine-tuning. The embedding size is 4096.
    \item \emph{Node2Vec.} Zhang et al.~\cite{zhang2023navigating} propose to leverage textual representations for HAR. More specifically, they pretrain label embeddings to align the feature space for HAR in distributed systems. They compute co-occurrence among label names in the BookCorpus dataset~\cite{lin2020ensemble} and learn semantic embeddings based on label similarity map~\cite{grover2016node2vec}.
\end{itemize}

\our and the compared methods are universal to all types of model architectures. To be effective, we test all methods on 1D ResNet~\cite{he2016deep} backbone and transformer~\cite{vaswani2017attention} backbone for they are the most representative architecture used by existing HAR methods~\cite{dempster2020rocket, li2021two, shavit2021boosting, kim2022radar}.
For each method, we experiment with the optimal combination of hyperparameters, including the choice of optimizer, learning rate, weight decay, and batch size. We reserve 10\% of the data as the testing set and further split 10\% of the training set as the validation set. The checkpoint with the best validation accuracy is adopted to calculate the testing accuracy. All experiments are run 5 times and the average result is reported. The standard deviation is not included, as most methods show consistent performance.

\begin{table*}[t]
\caption{Ablation Study on Selected Datasets. The average relative accuracy change is reported based on five runs.}
\begin{tabular}{ccccccc}
\toprule
Model & Representation
& MASD-IMU(H) & MASD-WiFi(E) & Opportunity & UTHAR & Mean \\ \midrule
\multirow{4}{*}{ResNet} 
& \our & 83.82 & 55.05 & 89.54 & 98.38 & 81.70 \\
& w/ coordinate recon & -2.86 & -7.30 & -1.04 & -1.13 & -3.08 \\
& \rev{w/ fine angle recon}
& \rev{-2.07} & \rev{-2.52}
& \rev{-0.68} & \rev{-1.16} & \rev{-1.61}
\\ 
& w/o attention matching  & -2.10 & -8.60 & -1.01 & -2.21 & -3.48 \\ \midrule
\multirow{4}{*}{Transformer} 
& \our & 81.86 & 65.57 & 88.75 & 98.12 & 82.00 \\
& w coordinate recon & -3.95 & +0.83 & -0.33 & -1.58 & -1.26 
\\
& \rev{w/ fine angle recon}
& \rev{-2.76} & \rev{-5.64} 
& \rev{-1.03} & \rev{-1.63}
& \rev{-2.77} \\
& w/o attention matching & -1.80 & -1.89 & -0.91 & -0.79 & -1.35\\ \bottomrule
\end{tabular}
\label{tab:ablation}
\end{table*}

The results in Table~\ref{tab:main_1} and Table~\ref{tab:main_2} show that \our outperforms all compared label representation methods. Although the improvement is around 2\%, this is substantial given that all the methods are using the same backbone models.
Since the label representation is obtained from a pretrained model, \our does not introduce extra model training efforts.
Furthermore, \our requires minimal modifications to the downstream HAR model structure, and incurs minimal extra computational costs. These advantages suggest that our method can be seamlessly integrated into existing HAR pipelines with accuracy improvements achieved at little to no additional cost.
One-hot representation is the longstanding 
choice and have competitive performance. Like the one-hot representation, the random vector treats labels independently but performs slightly worse than the one-hot. 
Llama-Emb performs well with the transformer on some datasets (e.g. MASD-IMU(H,M) and UTHAR), but underperforms on the other datasets. It mainly suffers from its high embedding dimensions which introduce redundant semantic information for HAR tasks.
Node2Vec performs well on the WiFi datasets but it is less effective on the IMU datasets. It is because Node2Vec is
trained on the BookCorpus, which captures general movement patterns. This is more similar to how WiFi sensing encompasses the entire body.

\subsection{Ablation Study}
We conduct an ablation study to analyze the effectiveness of the proposed techniques. Specifically, we compare two alternatives of coarse angle reconstruction for skeleton representation learning. We denote
\textbf{w/ coordinate recon} and \textbf{w/ fine angle recon} the objective of reconstructing joint coordinates and the fine-grained angles respectively. We also evaluate the attention matching and compare with \textbf{w/o attention matching} that simply matches HAR features with the label representations by averaging over the joints.

We follow the same settings in Section~\ref{sec:main} and experiment on
2 IMU and 2 WiFi datasets. The results are shown in Table~\ref{tab:ablation}.
In general, the ablation study suggests that both the coarse angle reconstruction loss and the attention matching module are helpful in improving the performance. The representations learned from other fine-grained objectives are not as informative as \our because they suffer from learning user-relevant information, which hurts their generalize-ability.
The only exception is that \textbf{w/ coordinate reconstruction} outperforms \our for MASD-WiFi(E) with Transformer. 
It suggests that coordinate reconstruction can help improve the performance but it requires the label set and the backbone model to fit the learned representation.
However, there is no concrete conclusion on how to apply it effectively.
Note that the improvement of the two mechanisms is less significant for the Opportunity dataset. It might be because it only has 4 easy locomotion activities (\textit{walking, standing, sitting, lying}) and the effect of a better representation method is limited for this dataset.

\subsection{Clustering Visualization}

\begin{figure*}[t]
\centering
    \includegraphics[width=0.93\textwidth]{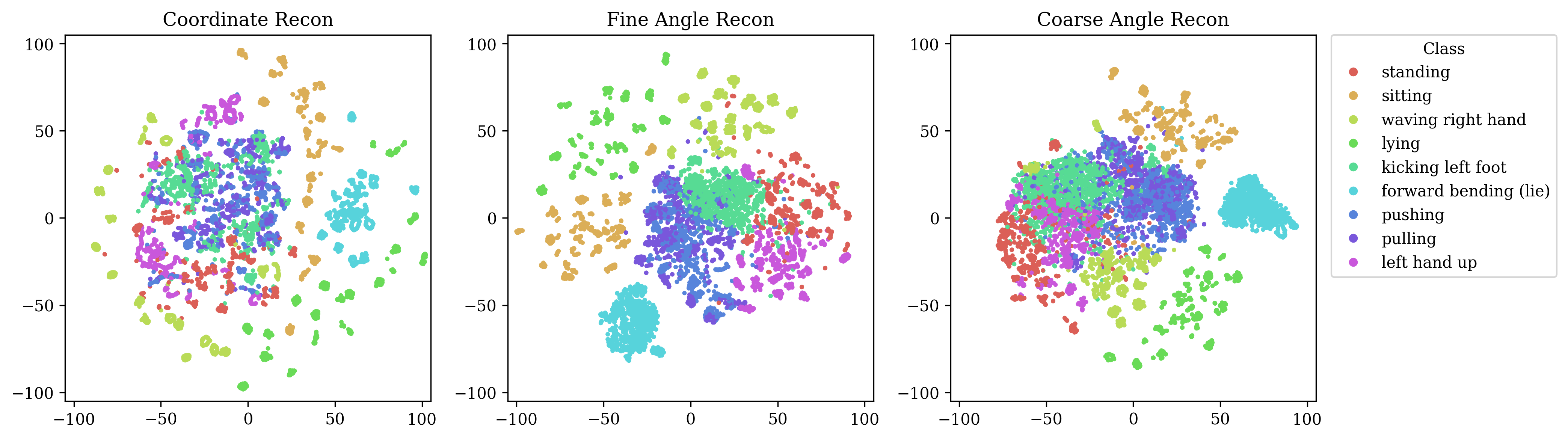}
    \caption{T-SNE of skeleton activity representation with colors annotating the \textit{activities}. Results indicate that angle-based reconstruction methods have a better clustering performance.} 
    \label{fig:tsne_1}
\end{figure*}

\begin{figure*}[t]
\centering
    \includegraphics[width=0.90\textwidth]{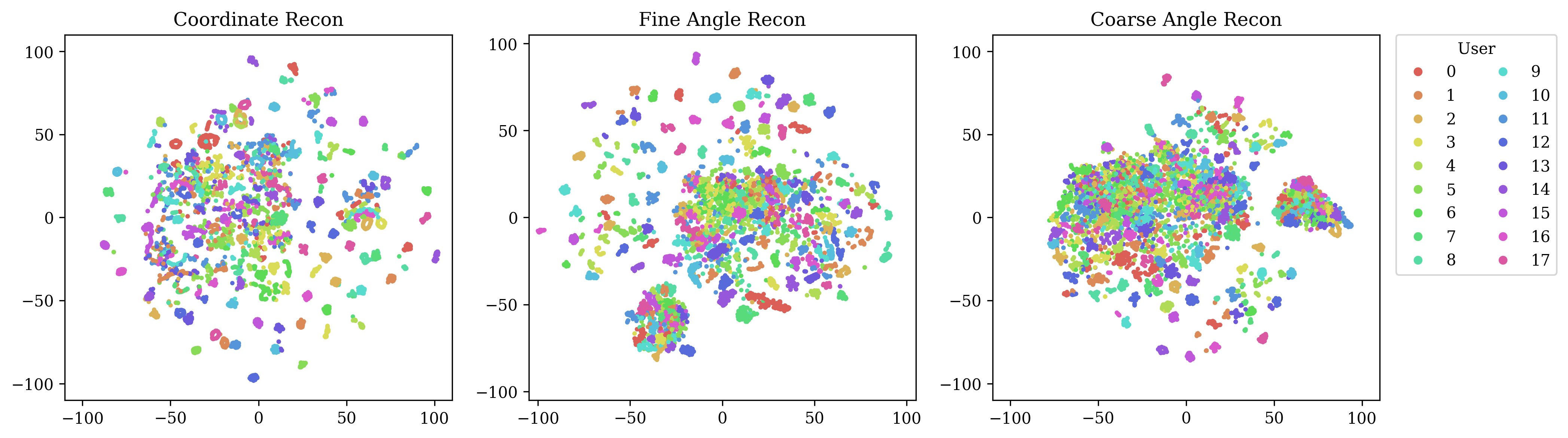}
    \caption{T-SNE of skeleton activity representation with colors annotating the \textit{users}. Results show that the coarse angle reconstruction is the most user-invariant, with fewest small cluster centers created.} 
    \label{fig:tsne_2}
\end{figure*}
We further compare the coarse angle reconstruction with the fine-grained angle, and the coordinate reconstruction using T-SNE to visualize the clustering performance in Figure~\ref{fig:tsne_1} and Figure~\ref{fig:tsne_2}.
Both figures demonstrate the same clustering results for a subset of activities in MASD. Note that we do not visualize the entire dataset because showing all the 27 activities can make the figure unperceivable. It is also very difficult for T-SNE to effectively map complicated skeleton representations to a simple 2D cluster space.
Figure~\ref{fig:tsne_1} marks the examples with the same \textit{label} using the same color, while  Figure~\ref{fig:tsne_2} marks the examples performed by the same \textit{user} using the same color. To reduce the difficulty of fitting the T-SNE model, we first take the average over the joints for dimension reduction.

\begin{figure}[t]
\centering 
    \includegraphics[width=0.47\textwidth]{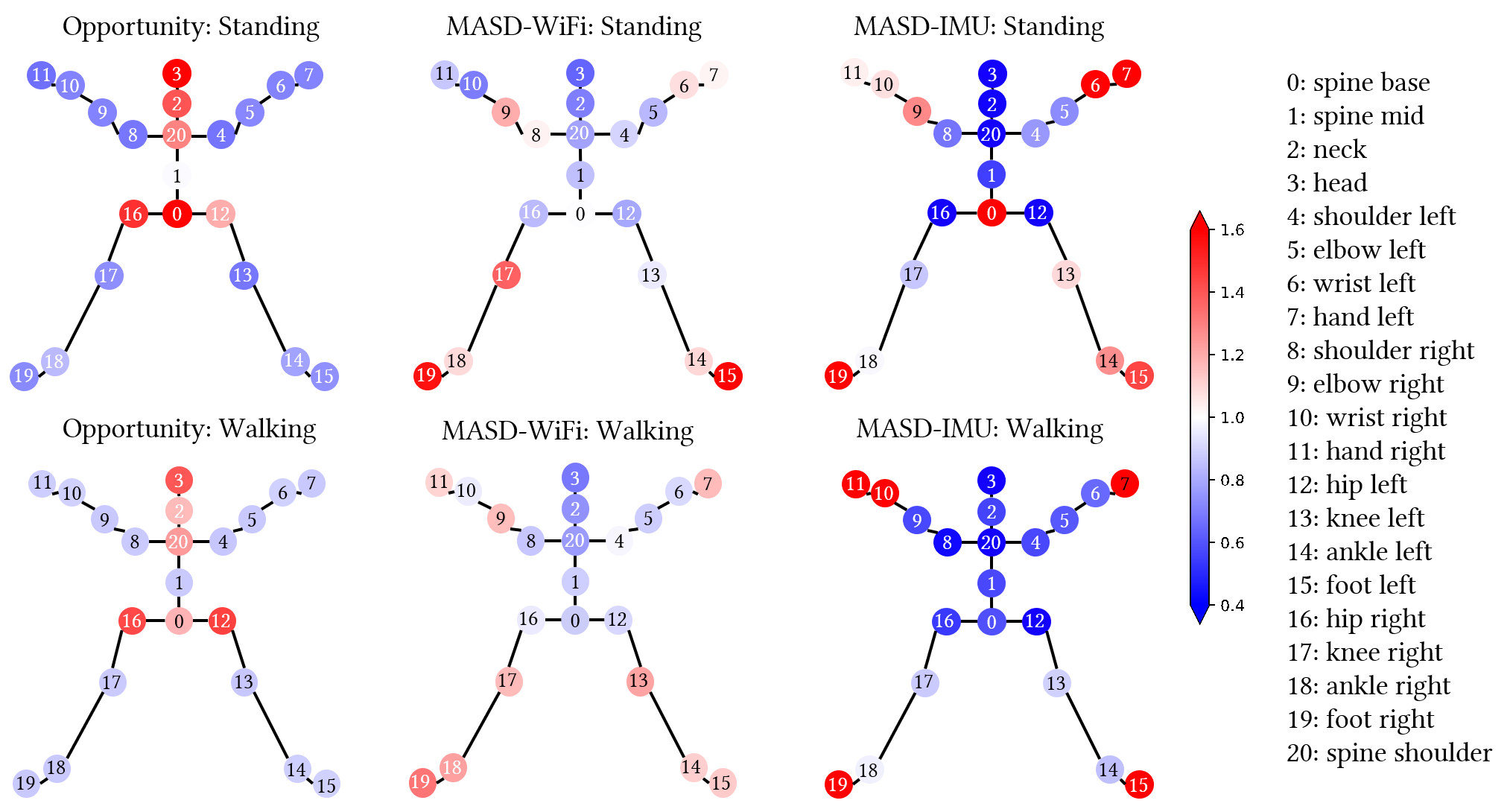}
    \caption{Visualization of the attention score on joints for ``standing'' and ``walking'' across three datasets. The warmer color indicates higher attention.} 
    \label{fig:attention}
\end{figure}

\begin{table*}[t]
\caption{5-shot Results on MASD Tasks. We report the average and the standard deviation over 5 runs. }
\begin{tabular}{cccccc}
\toprule

Model & Representation
& MASD-IMU(H) & MASD-IMU(M) & MASD-WiFi(E) &  Mean \\ \midrule

\multirow{5}{*}{ResNet} 
& \our w/o attention matching & \textbf{65.45 $\pm$ 1.72} & \textbf{73.17 $\pm$ 3.62} & \textbf{34.33 $\pm$ 5.47} & \textbf{57.65} \\
& One-hot & 60.17 $\pm$ 3.43 & 69.59 $\pm$ 6.15 & 32.96 $\pm$ 4.42 & 54.24 \\
& Random  & 64.04 $\pm$ 1.57 & 65.15 $\pm$ 6.78 & 32.96 $\pm$ 8.00 & 54.05 \\
& Llama-Emb & 63.03 $\pm$ 1.30 & 67.71 $\pm$ 3.52 & 27.30 $\pm$ 4.50 & 52.68 \\
& Node2Vec & 61.67 $\pm$ 4.43 & 68.53 $\pm$ 3.47 & 31.27 $\pm$ 9.55 & 53.82 \\ 
\midrule
\multirow{5}{*}{Transformer} 
& \our w/o attention matching  & \textbf{70.56 $\pm$ 2.32} & \textbf{76.99 $\pm$ 1.78} & \textbf{31.35 $\pm$ 1.06} & \textbf{59.63} \\
& One-hot   & 66.67 $\pm$ 1.82 & 75.61 $\pm$ 2.53 & 27.17 $\pm$ 1.44 & 56.48 \\
& Random  & 67.62 $\pm$ 2.43 & 74.42 $\pm$ 4.14 & 27.95 $\pm$ 3.52 & 56.63 \\
& Llama-Emb & 67.57 $\pm$ 3.21 & 75.33 $\pm$ 3.39  & 26.06 $\pm$ 5.54 & 56.32 \\
& Node2Vec  & 66.64 $\pm$ 1.55 & 75.08 $\pm$ 3.37 & 28.73 $\pm$ 10.21  & 56.82 \\ 
\bottomrule
\end{tabular}
\label{tab:few-shot}
\end{table*}

Figure~\ref{fig:tsne_1} shows that \our achieves better clustering compared to its fine-grained reconstruction counterparts.
The representations learned from coordinate reconstruction performs much worse than the two angle-based methods. Although it could distinguish activities like lying and sitting to some extent, it cannot generate clear cluster centers.
The two angle-based reconstructions can effectively group activities such as sitting, waving the right arm, lying, and bending. The clusters for most other labels are also well-defined, except for pushing and pulling, which are entangled. This overlap is expected, as pushing and pulling are highly similar activities involving arm stretching and expanding, and the only difference is the order of the movements. With a 5-second skeleton window size, samples often capture multiple rounds of these motions, which can lead to confusion for the skeleton encoder.
Standing and raising the right hand are also very close to each other, but they are clustered much better than the previous pair.

Comparing the angle-based reconstruction methods, coarse reconstruction produces more compacted cluster boundaries, with fewer small clusters scattered around the activity cluster centers.
We hypothesize that this is because the small clusters are actually user clusters. Fine-grained reconstruction is prone to learning these user-specific patterns, resulting in grouping the data from the same user.
To verify our conjecture, we shot Figure~\ref{fig:tsne_2} that demonstrates the same T-SNE result but marks the same user with the same color instead.
We observe that the small clusters are indeed the same users with the same activity type.
Coordinate reconstruction suffers greatly as the reconstruction objective does not explicitly decouple the two factors. Fine-grained angle reconstruction focuses on motion dynamics of the body skeleton, which mitigates the extraction of user-specific information. 
However, we still observe a clear trend that the learned representation effectively distinguishes between users.
Specifically, samples from the same users (0, 6, 10, 12, 13, 14) with different activities (pulling and pushing) is grouped very close to each other, which greatly harms the performance of fine-grained angle reconstruction for downstream HAR tasks.
We can also observe few small clusters in \our. But these clusters are still close to the activity centers, which demonstrates that the coarse angle reconstruction loss helps to learn more user-invariant features.

\subsection{Matching Attention Visualization}
We visualize the weights learned by the self-attention module on different sensing modalities in Figure~\ref{fig:attention}. We focus on \emph{standing} and \emph{walking} and demonstrate the self-attention score learned from the Opportunity, MASD-WiFi(E) and MASD-IMU(H) datasets.
Note that, the attention score is a reflection of both the sensing method and the characteristics of the activity type. 
We analyze Figure~\ref{fig:attention} from the two prospectives.

In terms of the sensing method, the attention for the Opportunity focuses on the torso, which corresponds to that one IMU sensor is deployed at the back of the users. 
WiFi sensing covers the entire body. As a result, the attention for the walking activity covers a wide range of joints in the arms and legs. MASD-IMU, on the other hand, only has IMU deployed at the right elbow and the left wrist. The attention of the upper body is also distributed in these locations. 

Considering activities types, the Opportunity focuses on the hip movement to recognize walking activities for these two joints are also close to the torso sensor. On the contrary, there is no sensor deployed at the lower body in MASD-IMU, thus the attention focuses on the feet parts because the movement of the feet can best summarize the characteristics of the walking activity. We also observe that there is little attention score on the arms for the Opportunity even though IMU sensors are actually deployed in these locations. It might be because, in the Opportunity dataset, users perform other signature-level activities (e.g. drinking water, eating or cleaning, etc.) while standing or walking. Focusing on the hands can introduce irrelevant information, thus misleading the model.

\begin{figure}[t]
\centering
    \begin{subfigure}[b]{0.22\textwidth}
        \includegraphics[width=\textwidth]{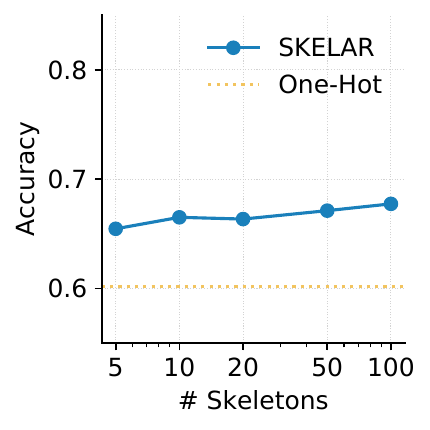}
        \caption{ResNet (Hard)}
    \end{subfigure}
    \begin{subfigure}[b]{0.22\textwidth}
        \includegraphics[width=\textwidth]{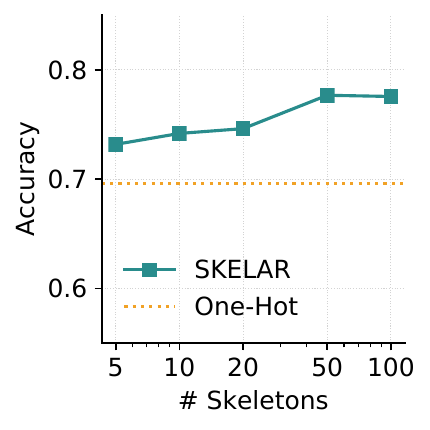}
        \caption{ResNet (Medium)}
    \end{subfigure}
    \begin{subfigure}[b]{0.22\textwidth}
        \includegraphics[width=\textwidth]{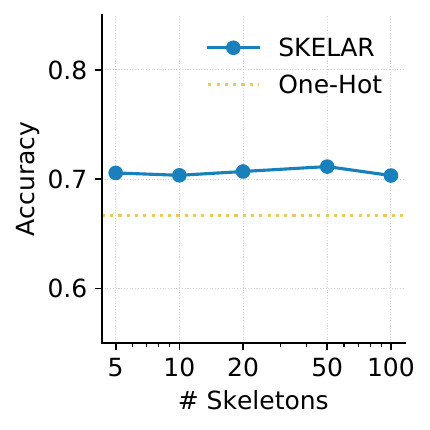}
        \caption{Transformer (Hard)}
    \end{subfigure}
    \begin{subfigure}[b]{0.22\textwidth}
        \includegraphics[width=\textwidth]{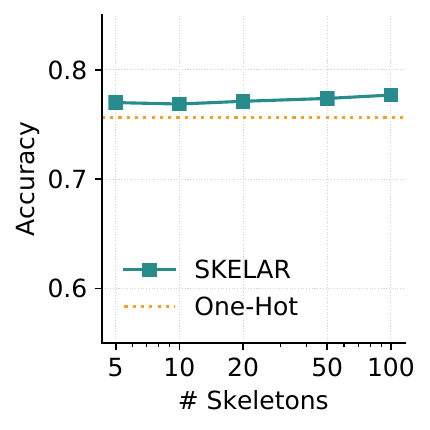}
        \caption{Transformer (Medium)}
    \end{subfigure}
    \caption{Few-shot accuracy on MASD-IMU (H, M) using ResNet and Transformer. We report the accuracy on 5 runs and the x-axis is plotted in log scale.} 
    \label{fig:exp-n-skeletons}
\end{figure}

\subsection{Few-Shot Analysis}
Few-shot learning is a practical scenario that trains models with a limited number of samples to achieve effective generalization~\cite{zhang2025contextual,zhang2023minimally,zhang2025react}.
Table~\ref{tab:few-shot} presents the 5-shot accuracy of \our and the baseline models on the MASD tasks. We report accuracy from the checkpoint with the lowest training loss. We further report the mean and the standard deviation of 5 runs using different random seeds since few-shot results can be quite inconsistent depending on the choice of the few-shot data. We test \our without the attention matching, as the self-attention is data-driven and it can severely overfit to the few-shot training samples. In general, we also outperform all the compared baselines and demonstrate a more significant improvement compared to the full-shot results. This is because the representations summarize the characteristics of the labels which can serve as classification criteria when labeled data is limited.

\our demonstrates a better improvement in the few-shot settings, which is mainly because it incorporates additional information learned from activity motion dynamics. 
We have also experimented on directly adapting SKELAR to a zero-shot framework by training the model on seen labels and testing on unseen labels. However, this is challenging as the model tends to overfit to the seen training labels~\cite{wang2023generalized}. A separate pipeline for matching embeddings of unseen labels would be required. Since this is beyond the scope of our paper, we leave it for future work.

\subsection{Sensitivity Analysis}
\label{sec:sensitivity}
We further study the sensitivity of \our to the number of skeleton samples used for acquiring the activity representation. Shown in Figure~\ref{fig:exp-n-skeletons}, we increase the number of skeleton samples from 5 to 100 and experiment with ResNet and Transformer models on MASD-IMU (H, M), together with the one-hot baseline noted by the dotted line. 
In this experiment, we use only 5 skeleton samples from each user, so the label representation is derived by averaging across all the 20 users.
Results show that with the increase of skeleton samples, \our can better represent activities and improve the performance. Note that the x-axis is in log scale, which means that the actual ascending trend in the curve is more significant.
This suggests that when the skeleton data is plentiful, acquiring the label representation from more users contributes to better consistency.
On the other hand, even with only 5 skeleton samples, we can still outperform the baseline with a significant margin, which emphasizes the flexible nature of \our as it only requires minimal labeled skeleton samples to obtain informative activity representations. 
\section{Discussion and Followup Challenges}
\label{sec:discussion}
\smallsection{Skeleton Encoder Structure.} Our skeleton encoder consists of multiple encoder blocks that capture spatio-temporal features at different resolutions. This design is driven by that various activities exhibit distinctive features across different temporal scales.
We have experimented with various skeleton encoder structures, including layer combinations and graph and temporal model architectures. However, this paper primarily focuses on using skeleton motion data as targets for downstream HAR tasks. We leave the design of a skeleton encoder for optimal performance in the future.

\smallsection{Broader Use Cases.}
\our can be directly extended to other modalities that also detect motion dynamics for activity sensing, such as radar and lidar.
For indirect sensing method such as vibration-based sensing, we need to calibrate the label representations to capture time-frequency domain information, enabling them to match with features extracted from vibration signal spectrograms.
Beyond full-shot HAR, \our has broader applications in label mismatching scenarios due to its flexibility to unseen activities.
For example, it can be used in distributed HAR systems to reconcile distinct label distributions from client devices.
Moreover, since our method captures comprehensive motion patterns, it can unify heterogeneous sensing systems that differ in deployment location, device type, and sensing modality,
enabling potential applications in multimodal HAR and cross sensing-domain adaptation.

\smallsection{Alternative Representations.}
The success of \our is based on the insight that skeleton data encodes high-level motion characteristics that can be generalized across diverse downstream sensing systems. 
We adopt a direct approach by learning label representations from the skeleton data and then adapting \our to various downstream HAR systems.
Another promising direction involves first adapting the skeleton data to the downstream system, followed by learning system-specific label representations.
Recent research has demonstrated the effectiveness of employing skeleton data to generate simulated IMU signals and then trains to align IMU time series with the textual description of labels~\cite{zhang2024unimts}. Therefore, applying this concept to other sensing modalities presents another exciting avenue for future work.

\smallsection{Edge Feasibility.}
\our can be deployed at the edge by maintaining a pretrained skeleton representation model on the cloud, with HAR matching performed on the edge devices.
To obtain new label representations, the edge device can communicate with the cloud server and request the model to generate the embeddings for unseen activity classes using the skeleton data collected in the lab environment or using virtually generated data. The label representations can be stored locally to avoid redundant communication overhead. 
The training of the edge model can take place locally, or adopt a heterogeneous federated learning setting~\cite{zhang2024few}. 
After model training, the attention layer can be entirely replaced with the cached attention-enhanced label representation. It further eliminates any additional computational cost during the inference phase.

\section{Summary and Conclusions}
In this paper, we propose \our, a novel framework that pretrains activity representation using skeleton-based motion data and then matches with heterogeneous sensing signals for HAR.
We design a coarse angle reconstruction self-supervised task that recovers the 3D rotation angles at essential body joints for pretraining the skeleton encoder.
We further propose a self-attention matching module that learns to focus on essential body joints in the downstream HAR tasks with heterogeneous sensing modalities and focuses.
To mitigate the lack of labeled skeleton data in existing HAR datasets, we present a new dataset, MASD with IMU, WiFi, and Kinect camera deployed, to evaluate the performance of \our using real-world skeleton activity. 
Moreover, we demonstrate that \our can support synthetic skeleton data to be quickly deployed in systems without skeleton collection needs as well.
Experiments show that our method achieves state-of-the-art performance in both full-shot and few-shot settings.
We further perform ablation studies and visualizations to understand the effectiveness of the two proposed mechanisms.

\begin{acks}
Our work is supported in part by ACE, one of the seven
centers in JUMP 2.0, a Semiconductor Research Corporation (SRC) program sponsored by DARPA and DataPlanet fellowship from Halicioglu Data Science Institute.

Our work is also sponsored in part by NSF CAREER Award 2239440, NSF Proto-OKN Award 2333790, as well as generous gifts from Google, Adobe, and Teradata. Any opinions, findings, and conclusions or recommendations expressed herein are those of the authors and should not be interpreted as necessarily representing the views, either expressed or implied, of the U.S. Government. The U.S. Government is authorized to reproduce and distribute reprints for government purposes not withstanding any copyright annotation hereon.

\end{acks}



\end{document}